\documentclass[letterpaper, 10 pt, conference]{formatting/ieeeconf}
\IEEEoverridecommandlockouts                              % This command is only needed if 
\let\labelindent\relax
\usepackage{amsfonts}       % An extended set of fonts for use in mathematics

\usepackage{amsthm}
\usepackage{amsmath}
\usepackage{mathtools}      % Loads amsmath
\usepackage{amssymb}        % provides AMS fonts and symbols ex. \mathbb{...}
\usepackage{filecontents}
\usepackage{graphicx}       % enhanced support for graphics
\usepackage{marginnote}     % Add notes
\usepackage{marvosym}       % mvat
\usepackage{overpic}        % Add text on figures
\usepackage{tabularx}
\usepackage{cite}
\usepackage{color}
\usepackage[linesnumbered,algoruled,boxed,lined]{algorithm2e}
\usepackage[normalem]{ulem}
\usepackage{epstopdf}
\usepackage{enumitem}
\usepackage[normalem]{ulem}
\usepackage{lipsum}
\usepackage{wrapfig}
\usepackage[a-2b,mathxmp]{pdfx}[2018/12/22]

\newif\ifdraft
\draftfalse

\ifdraft
\usepackage[paperheight=11in,paperwidth=9.5in,
			left=1.25in,right=1.25in,
			top=0.75in,bottom=0.75in,
			heightrounded,marginparwidth=1.2in,
			marginparsep=0.05in]{geometry}
\usepackage{xcolor}
\usepackage{xargs} 
\usepackage[textsize=footnotesize]{todonotes}
\newcommandx{\kg}[2][1=]{\todo[linecolor=red,
			backgroundcolor=red!10,bordercolor=red,#1]{#2}}
\newcommandx{\jy}[2][1=]{\todo[linecolor=green,
			backgroundcolor=green!10,bordercolor=green,#1]{JY:#2}}
   \newcommandx{\justin}[2][1=]{\todo[linecolor=orange,
			backgroundcolor=yellow!10,bordercolor=orange,#1]{Justin:#2}}
\else
\newcommand{\nt}[1]{{}}
\newcommand{\jy}[1]{{}}
\fi

\newif\iftwocolumn
\twocolumntrue

\newtheorem{problem}{Problem}

\theoremstyle{definition}

\theoremstyle{remark}

\DeclareMathOperator*{\argmax}{argmax}

\SetKwProg{Fn}{Function}{}{}
\SetKwComment{Comment}{$\triangleright$\ }{}

\makeatletter
\def\subsubsection{\@startsection{subsubsection}% name
                                 {3}% level
                                 {\z@ \hspace*{1mm}}% indent (formerly \parindent)
                                 {0ex plus 0.1ex minus 0.1ex}% before skip
                                 {0ex}% after skip
                                 {\normalfont\normalsize\itshape}}% style
\makeatother

\title{\bf%\LARGE \bf
Effectively Rearranging Heterogeneous Objects on Cluttered Tabletops 
}
\author{
Kai Gao \qquad Justin Yu \qquad Tanay Sandeep Punjabi \qquad Jingjin Yu
\thanks{
All authors are with the Department of Computer Science, 
Rutgers, the State University of New Jersey, Piscataway, NJ, 
USA. %E-Mails: \{{\tt kai.gao, jingjin.yu}\}\hspace*{.25em}
}%
\thanks{This work is partly supported by NSF awards IIS-1845888 and IIS-2132972, and an Amazon Research Award.
}
}

\def\toro{\texttt{TORO}\xspace}
\def\toroi{\texttt{TORI}\xspace}
\def\hete{\texttt{HeTORI}\xspace}
\def\heteCP{\texttt{HeCP}\xspace}
\def\heteTI{\texttt{HeTI}\xspace}

\begin{document}

\maketitle
\thispagestyle{empty}
\pagestyle{empty}

%%%%%%%%%%%%%%%%%%%%%%%%%%%%%%%%%%%%%%%%%%%%%%%%%%%%%%%%%%%%%%%%
% Latex editing related instructions for draft mode
%%%%%%%%%%%%%%%%%%%%%%%%%%%%%%%%%%%%%%%%%%%%%%%%%%%%%%%%%%%%%%%%
\ifdraft
\begin{picture}(0,0)%
\put(-12,105){
\framebox(505,40){\parbox{\dimexpr2\linewidth+\fboxsep-\fboxrule}{
\textcolor{blue}{
The file is formatted to look identical to the final compiled IEEE 
conference PDF, with additional margins added for making margin 
notes. Use $\backslash$todo$\{$...$\}$ for general side comments
and $\backslash$jy$\{$...$\}$ for JJ's comments. Set 
$\backslash$drafttrue to $\backslash$draftfalse to remove the 
formatting. 
}}}}
\end{picture}
\vspace*{-5mm}
\fi

%%%%%%%%%%%%%%%%%%%%%%%%%%%%%%%%%%%%%%%%%%%%%%%%%%%%%%%%%%%%%%%%
% Main text
%%%%%%%%%%%%%%%%%%%%%%%%%%%%%%%%%%%%%%%%%%%%%%%%%%%%%%%%%%%%%%%%
\begin{abstract}
Effectively rearranging heterogeneous objects constitutes a high-utility skill that an intelligent robot should master. Whereas significant work has been devoted to the grasp synthesis of heterogeneous objects, little attention has been given to the planning for sequentially manipulating such objects.
In this work, we examine the long-horizon sequential rearrangement of heterogeneous objects in a tabletop setting, addressing not just generating feasible plans but near-optimal ones. 
Toward that end, and building on previous methods, including combinatorial algorithms and Monte Carlo tree search-based solutions, we develop state-of-the-art solvers for optimizing two practical objective functions considering key object properties such as size and weight.
Thorough simulation studies show that our methods provide significant advantages in handling challenging heterogeneous object rearrangement problems, especially in cluttered settings.
Real robot experiments further demonstrate and confirm these advantages. 

\vspace{1mm}
%\noindent Code (upon publication): \url{https://youtu.be/BlahBlah}
\noindent Source code and evaluation data associated with this research will be available at \url{https://github.com/arc-l/TRLB} upon the publication of this manuscript. 
%\noindent Simulation video: %\url{https://youtu.be/BlahBlah}
\end{abstract}
%\justin{don't forget to change this}

\section{Introduction}\label{sec:intro}
In pick-n-place-based multi-object rearrangement problems, a robot is tasked to move objects, generally one at a time, from a start arrangement to a desired goal arrangement.
Multi-object rearrangement is a fundamental robotic task in commercial, domestic, and many other settings.
In supermarkets, a robot can assist in maintaining organized shelves.
At home, a robot can be utilized for tasks such as cleaning up kids' rooms or reorganizing a messy desk.

\begin{figure}[t]
\vspace{2mm}
    \centering
    \includegraphics[width=\columnwidth]{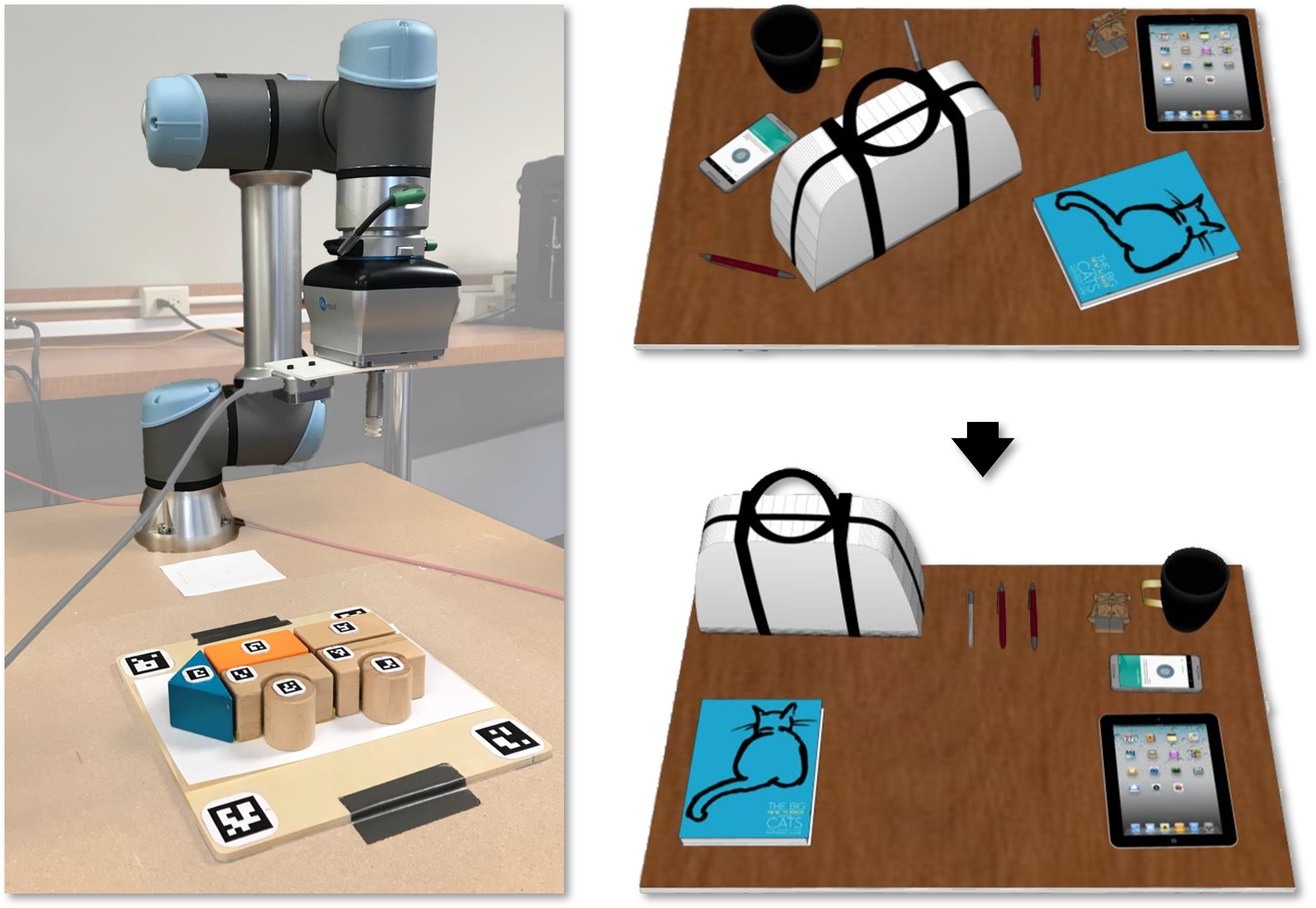}
    \caption{[Left] Our hardware setup for evaluating  tabletop rearrangement of multiple heterogeneous objects. One of the test cases is also shown. [Right] The illustration of an example instance where many objects of drastically different sizes and weights are to be rearrangements.}
    \label{fig:intro_example}
\end{figure}

Among rearrangement problems, Tabletop Object Rearrangement with Overhand grasps (\toro) has been extensively studied due to its many applications to real-world scenarios. The apparently simple setup is in fact NP-hard to optimally solve\cite{han2018complexity}. 
The difficulty is due to the overlaps between the start and goal arrangements, which makes it impossible to directly move some objects to the goal poses before some other objects are first moved away.
For example, in Fig.~\ref{fig:intro_example}[Right], the white bag cannot move to the goal pose before the black mug moves away from the current pose.
When many such dependencies exist, they entangle and give rise to a combinatorial challenge. 
Deciding the best sequence to untangle the dependencies and minimize the total number of temporarily displaced objects makes the problem NP-hard.  
%

%As a result, \toro is a combinatorial challenge requiring the robot to find a feasible/optimal sequence of object movements.
In addition to the combinatorial challenge of minimizing the total number of displaced objects, another potentially more thorny issue in solving \toro is finding ideal locations (i.e., \emph{buffers}) for temporarily placing objects that cannot be directly moved to their goals. 
This issue appears when the workspace is cluttered with limited free space.  
%
%as some objects must be placed in temporary locations before moved to their goal, when the amount of free space is limitied some objects need temporary displacements (``buffer'') in free space.
In the example from Fig.~\ref{fig:intro_example}[Right], the book and the phone cannot move to the goal pose before the other object is moved away, so at least one of these two objects needs to be moved to a \emph{buffer} to break the mutual dependency. 
Suppose we are limited to using the tabletop as the workspace, it is not immediately clear which object should be moved away and where within the tabletop the moved object should be temporarily displaced. 
We denote this subclass of \toro as \toroi (a \toro in which temporarily displaced objects must remain \emph{internal} to the tabletop workspace) \cite{gao2022fast}.

Until now, algorithmic research on \toroi is limited to solving instances with mostly homogeneous objects \cite{han2018complexity,labbe2020monte,shome2020fast,
%gao2021running,
gao2022fast
%,gao2022toward
}. 
%\jy{Kai: please provide relevant references, including our work and some of Kostas', and maybe more.}
%
The assumption of homogeneous objects leads to a relatively simple model where each object manipulation action has the same or very similar cost, which in turn limits the applicability of the resulting methods. 
In many real-world tabletop rearrangement problems, objects are often heterogeneous. For example, in Fig.~\ref{fig:intro_example}, some objects are small and thin, e.g., pens, while others can be bulky, e.g., the white bag. Intuitively, objects' sizes and shapes impact the rearrangement planning process - it is more difficult to relocate larger objects temporarily. 
Furthermore, other properties, such as weight, affect the effort needed to move them around. This further complicates the computation of optimal rearrangement plans.  
% Intuitively speaking, moving large objects or stick-like objects are more likely to make the goal poses of other objects available due to the overlap between their footprints and those of other objects in goal poses.
% However, it is also more difficult to allocate buffer locations for these objects and their goal poses are more likely to be occupied by more other objects.
We thus ask a natural question here: 
% How much impact do object size and shape have on the difficulty of buffer allocation? And 
Can we leverage objects' properties to guide multi-object rearrangement planning?

With the question in mind, we investigate heterogeneous \toroi (\hete), in which the characteristics of workspace objects, such as size, shape, weight, and other properties, can vary significantly. 
Taking object characteristics into consideration immediately leads to a much more general problem model, allowing each manipulation operation to have a cost that is affected by the properties of the objects. 
For example, manipulating an object that is large, heavy, and/or fragile is naturally more challenging. As a result, such an operation could take more time to complete. Therefore, a higher cost may be given for such operations than manipulating an object that is small, light, and/or tough. 
This work focuses on developing methods taking into account such differences. 
More specifically, we develop methods for addressing two problem variants. In one of the variants, object shape and size are considered in generating the manipulation plan. Object characteristics are directly reflected in the optimization objective in the other variant.

%\jy{My edit location. Need to describe the two methods that are developed: one that does not assign different costs to different operations and one that does.}
%\jy{For contribution, we should highlight two: (1) We tackle the problem where object sizes can be drastically different in a cluttered setting, and introduce a more precise model for TORO, (2) We studied two sub-models where one does not differentiate manipulation cost and one that does. Our methods are effective. }
%We can use object properties such as mass or size as ``impedance'' of manipulating each object in the task space and seek the rearrangement plan minimizing the total task impedance.

%
%First, object characteristics can be used as prior knowledge to guide the planning process. 
%Although large or stick-like objects may be difficult to relocate, a planner with size and shape information can find feasible plans more efficiently.
%Second, object property information allows us to solve the rearrangement problem with higher-fidelity objectives.

In studying \hete, we deliver the following results and contributions: 
\begin{itemize}[leftmargin=3.5mm]
    \item The formulation and investigation of the \hete problem, to our knowledge, is a first in introducing significant object heterogeneity in long-horizon manipulation task planning, adding not only generality but also additional computational complexity. In contrast, previous object rearrangement research generally assumes uniform objects and/or uniform manipulation costs. 
    \item We examine \emph{optimally} solving \hete over two practical objectives, with one assuming uniform manipulation costs but taking into account object heterogeneity and the other directly associating manipulation costs with object heterogeneity, i.e., larger objects take longer to move around. For both objectives, we develop SOTA methods based on deterministic and Monte Carlo tree search that greatly enhance computational efficiency and solution optimality, particularly for cluttered settings.  
    \item Thorough simulation studies show that considering object heterogeneity translates into better plans for solving \hete, compared with approaches ignoring individual objects' differences. Real robot experiments further confirm these findings.   
\end{itemize}

\textbf{Paper Organization} The rest of the paper is organized as follows. 
We first provide an overview of previous works in related fields in Sec.~\ref{sec:related}.
In Sec.~\ref{sec:problem}, 
we formally present the studied problem \hete. 
Then, in Sec.~\ref{sec:method}, we describe algorithmic approaches used to solve \hete.
In Sec.~\ref{sec:evaluation}, we evaluate the performance of our proposed methods.
Finally, we present our conclusion in Sec.~\ref{sec:conclusion}.

\section{Related Work}\label{sec:related}
%\textbf{Object Manipulation and Rearrangement}
Robotic manipulation \cite{9893496} can be categorized into prehensile (e.g., \cite{wang2022efficient, gao2022toward
%, gao2022utility
}) or non-prehensile (e.g., \cite{huang2019large,han2021toward,song2020multi,huang2021dipn}) methods.
Non-prehensile manipulations, e.g., pushes or pokes, are simple to execute but less predictable. Prehensile manipulations, on the other hand, result in more precise movements that facilitate long-horizon planning.
Object rearrangement is widely applicable to diverse scenarios. The most typical ones are tabletop\cite{gao2022toward,danielczuk2021object,shome2020fast} and shelf setups \cite{wang2022lazy,wang2022efficient,wada2022reorientbot} with fixed robot arms. 
%\jy{Kai: please provide relevant references, e.g., Thanasis' and Rui's work.}
%
Larger environments with mobile robots\cite{gu2022multi,gan2022threedworld,szot2021habitat} have also been examined.
Typical rearrangement problems seek plans to move objects to desired goal positions, some variations, e.g., NAMO (Navigation Among Movable Obstacles)\cite{stilman2005navigation
%,stilman2008planning
} or object retrieval problems\cite{huang2021visual,ahn2022coordination,vieira2022persistent,vieira2022persistentmcts}, clear out a path to relocate a target object or the robot among movable obstacles.
% Earlier works study NAMO (Navigation Among Movable Obstacles), which clears out a path to a target location in a cluttered environment with movable obstacles.
% Similar to NAMO, object retrieval problems seek feasible plans to retrieve objects from a shelf or on a cluttered tabletop.

In \toro, object collisions due to overlaps between the start and goal arrangements can be encoded into a dependency graph, which connects the rearrangement problem to established graph problems\cite{han2018complexity,gao2021running,gao2023minimizing}.
With the dependency graph, efficient algorithms and structural properties have been obtained for \toro.
Specifically, assuming external space for temporary placements, the challenge of \toro embeds NP-hard Feedback Vertex Set (FVS) problem and Traveling Salesperson Problem (TSP)\cite{han2018complexity}.

Our previous work\cite{gao2021running} studies another natural objective for \toro, minimizing the number of \emph{running buffers}, which is the number of currently displaced objects (in some space external to the workspace).
It turns out that the running buffer objective can effectively help free space management in cluttered workspaces, leading to a speedy and high-quality method, TRLB (Tabletop Rearrangement with Lazy Buffers),  for solving \toro problems \cite{gao2022fast,gao2023minimizing}.
%Inside TRLB (Tabletop Rearrangement with Lazy Buffers)\cite{gao2022fast}, a framework for \toro without external space, the idea of minimizing running buffers effectively helps free space management in cluttered workspaces.
Besides dependency graph-based methods, there are some recent works on \toro utilizing reinforcement learning models\cite{yuan2018rearrangement,yuan2019end,bai2021hierarchical,kumar2022graph} and Monte Carlo Tree Search (MCTS)\cite{labbe2020monte}.
In examining \toro with heterogeneous objects, this work boosts the performance of the dependency graph-based method TRLB and the MCTS-based method with the aid of object characteristics.

%\textbf{Manipulation with Heterogeneous Objects}
%
The study of manipulating heterogeneous objects is generally limited to the grasp synthesis of single objects \cite{bicchi2000robotic,bohg2013data, sahbani2012overview}.
With the aid of powerful simulators, recent works synthesize grasps by ranking sampled candidate grasps\cite{przybylski2011planning} or using data-driven methods \cite{newbury2022deep,morrison2018closing}.
In addition to the challenge of searching for reachable grasps, for unknown objects, grasp poses are sampled based on local or global shape features identified from the sensing data\cite{wang2020feature,fischinger2013learning}.
To our knowledge, sequential manipulation planning of heterogeneous objects has received little attention.
%However, instead of a sequence of actions, the task in these papers tends to be a single manipulation, such as a grasp, and a pick-n-place.
%To the best of our knowledge, research about the impact of object shape and size on task planning is scarce.
In this research, we investigate effective ways to leverage object properties to guide the rearrangement planning of heterogeneous objects.

\section{Preliminaries}\label{sec:problem}
In this section, we formally define \hete.  We then explain the dependency graph structure and Tabletop Rearrangement with Lazy Buffers (TRLB), a method for \toroi. An alternative method \cite{labbe2020monte} based on Monte Carlo Tree Search \cite{kocsis2006bandit} is also briefly explained.

\subsection{Heterogeneous \toroi (\hete)}\label{sec:hete-tori}
Consider a bounded rectangular tabletop workspace $\mathcal W \subset \mathbb{R}^2$ in front of a robot arm.
The robot arm can reach every point on $\mathcal W$.
There is a set of $n$ objects $\mathcal O=\{o_1, o_2, \dots, o_n\}$ in $\mathcal W$.
Each object $o_i$, $1\le i\le n$, is an upright generalized cylinder. The generalized cylinder assumption, together with the overhand grasp/release, implies that the effective object pose space is $SE(2)$.
A pose $p_i=(x,y,\theta)\in SE(2)$ for object $o_i$ is \emph{valid} if $o_i$ at $p_i$ is inside $\mathcal W$ and does not collide with other objects at their current poses.
Otherwise, we say pose $p_i$ is \emph{in collision} (with some pose $p_j$ for some object $o_j$).
An object arrangement $\mathcal A=\{p_1,p_2,\dots,p_n\}$ is \emph{feasible} if all the poses in $\mathcal A$ are valid poses.
We consider overhand pick-n-places for object manipulation.
Each pick-n-place $a$ can be represented as $(o_i, p_i)$,  where the robot arm $R$ grasps $o_i$ at the current pose from above, lifts it up, moves it horizontally over a valid pose $p_i$, and finally places it down.
The task is to compute a \emph{rearrangement plan}, which is a sequence of pick-n-places $\Pi=\{a_1, a_2, \dots, a_{|\Pi|}\}$ moving objects one at a time from a feasible start arrangement $\mathcal A_s$ to a feasible goal arrangement $\mathcal A_g$.

\hete assumes full knowledge of object characteristics $\mathcal I=\{I_1, I_2,\dots,I_n\}$.
$\mathcal I$ may include properties of objects, including geometry, mass, material properties, and so on. 
In practice, $\mathcal I$ can be estimated based on sensing data.

% For each pick-n-place, the execution time contains the traveling time for horizontal moves in the workspace and manipulation time. 
On a tabletop, a pick-n-place action consists of object grasp/release and end-effector travel. In the vanilla version of \toro, it is typically assumed that the time required for the end-effector to move is negligible compared to the time needed for grasping and releasing objects.
%the traveling time for horizontal moves, the manipulation time for grasping and placing tends to dominate the execution time due to the needed accuracy for grasping/placing and the relatively small scale of the tabletop workspace.
Therefore, the quality of rearrangement plans can be evaluated based on the total number of grasps (and an equal number of releases) in solving a task. 
For \hete, if the robot is sufficiently powerful, it may still be the case that grasping/releasing objects takes the same amount of time for each object, regardless of their sizes and mass (with some range); but it could also be that different objects requires different (time) cost to grasp/release.
Based on the observation, for \hete, two cost objectives are examined in this study, which are: 
\begin{enumerate}
    \item \textbf{(\texttt{PP})} Total number of relocations, i.e., $J(\Pi)=|\Pi|$.
    \item \textbf{(\texttt{TI})} Total task \emph{impedance}, i.e., \vspace{2mm}\begin{align}J(\Pi)=\sum_{(o_i,p_i^1,p_i^2)\in \Pi} f(I_i).\label{eq:ti}\end{align}
\end{enumerate}

Specifically, the PP objective is the number of grasps or places a robot needs to execute in a rearrangement plan.
The TI objective enables the prioritization of manipulation actions based on objects' intrinsic properties, where $f(I_i)$ represents the actual cost of manipulating object $o_i$, as determined by the robot and $I_i$. 
Compared to PP, the TI objective represents the manipulation cost with higher levels of fidelity. 
For example, if a robot is tasked to rearrange gold and iron bars of the same size and shape, it is certainly better to move iron bars more due to iron's significantly lower density.  
%This feature is useful in \hete since one may want to frequently pick and place lightweight or handy objects but avoid manipulations of heavy or fragile objects.
%$f(I_i)$ measures the cost of moving $o_i$ based on information $I_i$. 
%It can be the mass of object $o_i$ or the estimated difficulty of grasping $o_i$.

Based on the descriptions so far, \hete can be formally summarized as follows:

\begin{problem}[\hete]
Given two feasible arrangements $A_s$ and $\mathcal A_g$ of objects $\mathcal O$, compute a rearrangement plan $\Pi$ moving $\mathcal O$ from $\mathcal A_s$ to $\mathcal A_g$ with the minimum cost $J(\Pi)$.
\end{problem}

\subsection{Dependency Graph}\label{sec:dg}
\begin{figure}[ht]
    \centering
    \vspace{1mm}
    \includegraphics[width=0.2\textwidth]{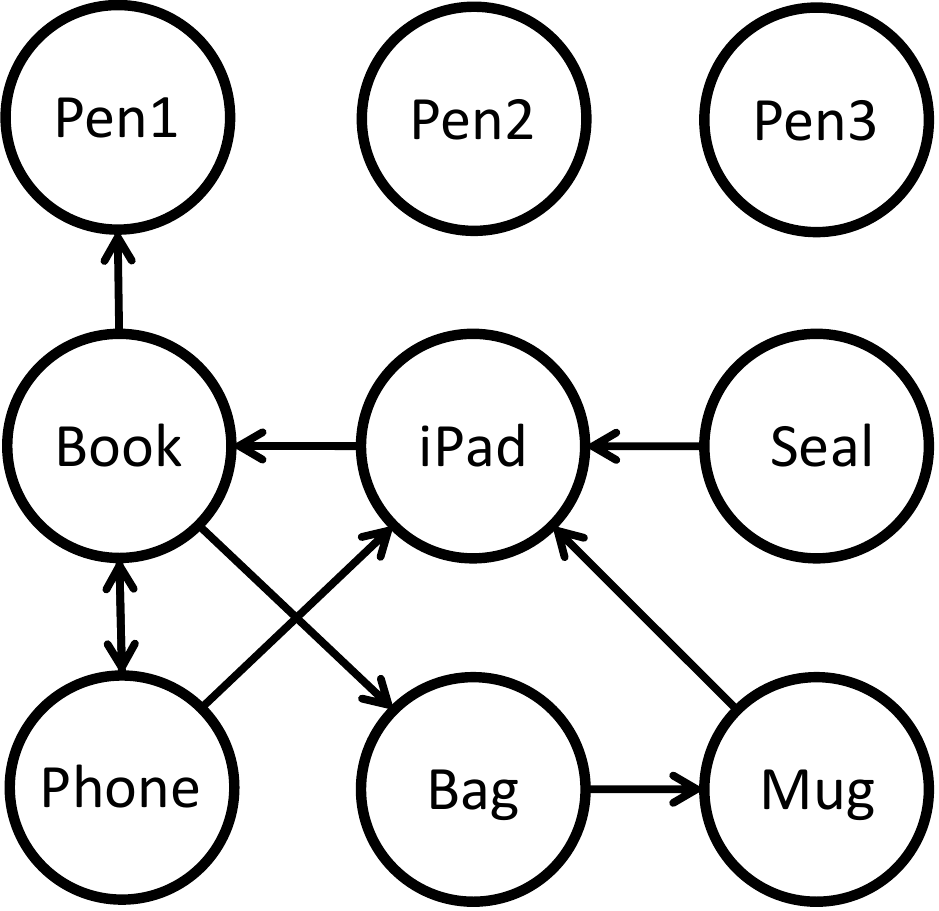}
    \vspace{-1mm}
    \caption{The corresponding dependency graph of for the \toro (and also a \hete) instance from Fig.~\ref{fig:intro_example}.}
    \label{fig:dp_example}
\end{figure}

In \toro, due to pose collisions between the start and goal arrangements, some objects cannot be moved to goal poses before others are moved away.
This leads to \emph{dependencies} among objects.
Specifically, we say an object $o_i$ \emph{depends on} another object $o_j$ if the goal pose of $o_i$ intersects (collides with) the start pose of $o_j$.
Based on object dependencies, we construct a \emph{dependency graph} $\mathcal G$ to guide the task planning process for in \toro.
In $\mathcal G$, each vertex $v_i$ represents an object $o_i$, and there is an arc from $v_i$ to $v_j$ if $o_i$ depends on $o_j$.
$\mathcal G$ encodes pose collisions between the start and goal arrangements.
Fig.~\ref{fig:dp_example} shows the corresponding dependency graph of the \toro instance in Fig.~\ref{fig:intro_example}.

When a dependency graph is acyclic, objects can be moved directly to the goal poses following the topological order.
When the dependency graph contains cycles, some objects must be relocated to ``break the cycles'' before other objects in the cycles and themselves can be moved to the goal poses.
The benefits of using the dependency graph $\mathcal G$ are threefold:
First, $\mathcal G$ contains the information for computing the least number of object relocations to complete the rearrangement.
For example, in Fig.~\ref{fig:dp_example}, removing the book vertex from $\mathcal G$ breaks all three cycles in $\mathcal G$, which suggests temporarily displacing the book to a collision-free buffer pose.
Second, the strongly connected components in $\mathcal G$ help decouple the rearrangement problem.
In Fig.~\ref{fig:dp_example}, there is one 5-vertex component and four single-vertex components.
We can solve the subproblem induced by each component individually, which helps reduce the problem scale when dealing with many objects.
Third, $\mathcal G$ pre-computes collisions between start and goal arrangements. 
In general rearrangement problems, at each timestep, most of the objects are either at start poses or goal poses.
Therefore, referring to $\mathcal G$ saves much time on collision detection efforts.

%\justin{nitpick: "iPad"}

\subsection{Tabletop Rearrangement with Lazy Buffers (TRLB)}\label{sec:trlb}
The dependency graph $\mathcal G$ suggests a decoupling of the rearrangement problem and a set of objects moved to buffers but it cannot suggest collision-free buffer poses for these objects inside the workspace. 
To solve \toroi, TRLB (Tabletop Rearrangement with Lazy Buffers) \cite{gao2022fast} uses $\mathcal G$ to guide the rearrangement planning process.
% TORI can be efficiently solved by delaying the buffer allocation process, i.e. tabletop rearrangement with lazy buffers (TRLB).
TRLB computes a rearrangement plan in two steps. 
First, it computes a sequence of primitive actions, each of which indicates an object to move to either a buffer pose or the corresponding goal pose.
In the example of Fig.~\ref{fig:intro_example}, a primitive action can be a command like ``move the book to a buffer'' or ``move Pen3 to the goal''.
After that, in the second step, for objects that need buffers, TRLB searches for buffer poses not colliding with other objects or blocking later primitive actions.
When the two-step strategy fails, TRLB records a partial rearrangement plan and conducts a bi-directional search to concatenate partial plans into a complete plan.
TRLB efficiently computes high-quality plans and is shown to be effective in solving dense instances when objects are homogeneous.

For primitive plan computation, TRLB provides two options: TBM (total buffer minimization) and RBM (running buffer minimization).
TRLB with TBM computes primitive plans minimizing the total number of primitive actions moving objects to the buffers.
TBM is equivalent to computing the minimum feedback vertex set of the dependency graph $\mathcal G$, a set of vertices in $\mathcal G$ whose removal makes $\mathcal G$ acyclic.
The optimal solution of TBM can be obtained via dynamic programming or integer programming\cite{han2018complexity, gao2021running}.
This option seeks the shortest primitive plan but is not scalable to the number of objects and workspace density levels (how cluttered the workspace is) due to the computational complexity and the neglect of the difficulty of buffer allocation.
TRLB with RBM computes primitive plans minimizing the maximum number of objects concurrently staying at buffer poses.
RBM can be computed by DFDP (depth-first dynamic programming), a method of performing exhaustive state space search in a depth-first manner.
It is a more efficient alternative to TBM but may return slightly longer primitive plans.

\subsection{MCTS for Solving \toroi}
%\jy{Kai: please add a brief description of the MCTS method.}
Labbe et. al.\cite{labbe2020monte} propose a Monte Carlo Tree Search (MCTS) based method  for \toroi.
Each state in MCTS represents an arrangement of objects.
At each state, an action $a_i$ either moves object $o_i$ to the goal pose or relocates an object blocking $o_i$'s goal pose to a collision-free pose.
The reward is the number of objects at goals.
MCTS selects actions based on the upper confidence bound (UCB):
\begin{equation}\label{eq:ucb}
    \argmax_a(
    \dfrac{\omega(f(s,a))}{n(f(s,a))}
    ) + C\sqrt{\dfrac{2\log(n(s))}{n(f(s,a))}},
\end{equation}
where $f(s,a)$ is the resulting state from state $s$ after an action $a$, $n(s)$ and $\omega(s)$ represent the number of times that state is visited and the total rewards after the visits of $s$. 
In Eq.\ref{eq:ucb}, the first term is an exploration term, representing the expected rewards after action $a$, and the second term is the exploitation term, prioritizing the actions with fewer visits.

\section{Methodology}\label{sec:method}
Our algorithmic design methodology is general: different (weighting) heuristics can be employed by different seach strategies (e.g., TRLB and MCTS) for solving \hete. 
Here, we first describe our carefully constructed heuristics that are applicable to both PP and TI objectives.
%
%Our approach is general: weighting heuristics for additional objective functions can be proposed similarly.
Then, for TRLB, we construct a weighted dependency graph with the heuristics and formulate the corresponding feedback vertex set problem and running buffer minimization problem.
For MCTS, we use the heuristics with additional improvements.

\subsection{Weighting Heuristics}\label{sec:weighting}

\subsubsection{Collision Probability  (\heteCP) for PP}\label{sec:collision}
In a cluttered workspace, it is difficult to allocate collision-free poses for use as buffers.
Even when a pose is collision-free in the current arrangement, it may overlap/collide with some goal poses, blocking the movements of other objects.
It is generally more challenging to allocate high-quality buffers for objects with large sizes or aspect ratios (e.g., stick-like objects).
A natural question would be:
How much do the object's size and shape impact its collision probability with other objects? 
We compute an estimated collision probability for each object to answer the question. Based on the collision probability, we propose a heuristic \heteCP (Heterogeneous object Collision Probability) to address the buffer allocation challenge in \hete and guide the rearrangement search. 
% \heteCP suggests the difficulty of allocating high-quality buffers.

Given two objects $A$ and $B$ with fixed orientations, positions of $B$ colliding with $A$ can be represented by the \emph{Minkowski difference} of the two objects. That is,
$$
A\ominus B(0,0) := \{a-b| a \in P_A, b\in P_B(0,0)\},
$$
where $P_A$ is the point set of $A$ at the current pose and $P_B$ is the point set of $B$ at position $(0,0)$ with the current orientation. 
Note that when $B$ is rotational symmetric, the Minkowski difference is the same as the Minkowski sum, which replaces the minus with a plus in the formula above.
%\jy{Please update to using Minkowski sum.}

\begin{wrapfigure}[10]{r}{1.5in}
    \centering
    \vspace{-2mm}
    \includegraphics[width=1.5in]{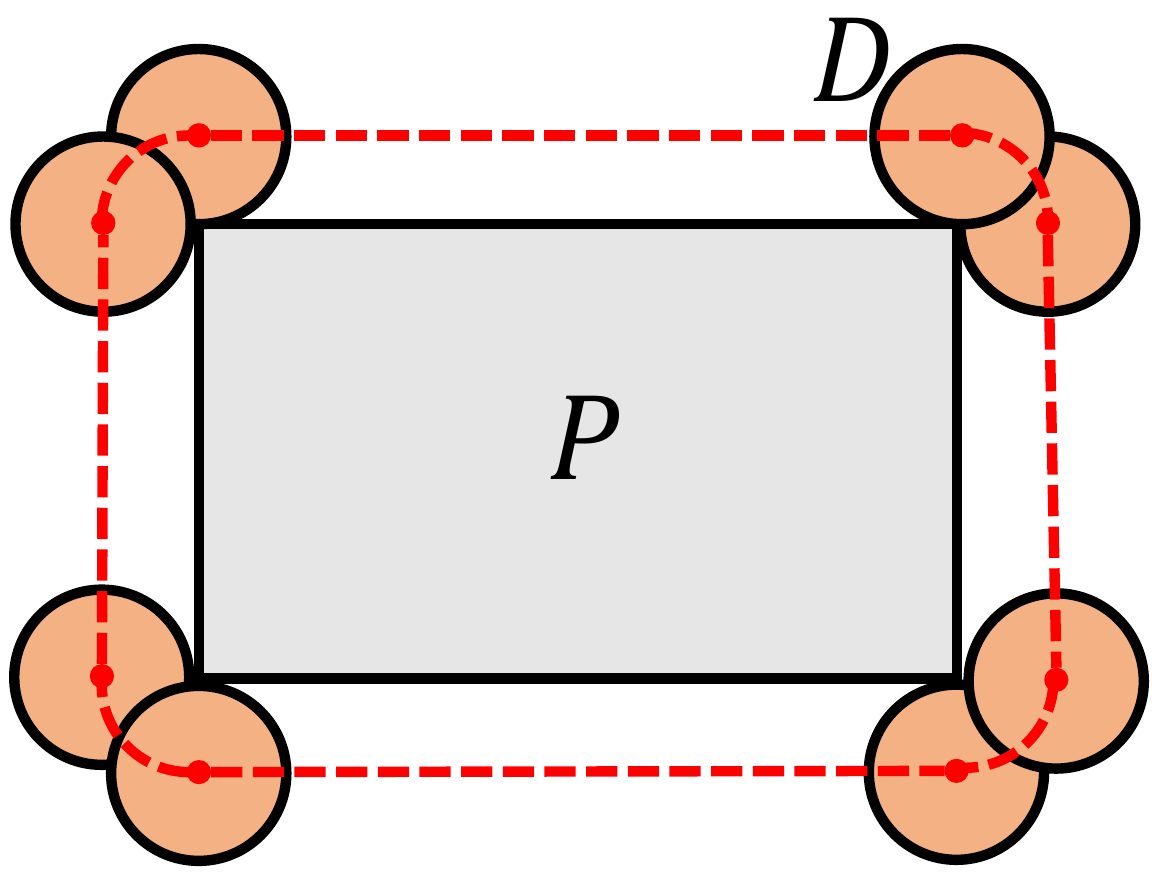}
    \vspace{-6.5mm}
    \caption{An illustration of the Minkowski difference $P\ominus D(0,0)$ (red dashed region) of a rectangle $P$ and a disc $D$.}
    \label{fig:no_fit}
\end{wrapfigure}

Specifically, for a convex polygon $P$ and a disc $D$, the area of $P\ominus D(0,0)$ can be simply represented as
$
S_{N}=S_{P}+S_{D}+r_DC_{P},
$
where $S$ and $C$ are the area and circumference of corresponding objects, $r_D$ is the radius of $D$. 
In the example of Fig.~\ref{fig:no_fit}, $S_P$ comes from the polygon $P$ in the center of $N$, $S_D$ comes from the four rounded corners, and $r_DC_P$ comes from the area along the edges.
Since one of the objects is a disc, the formula is rotation invariant. 
Furthermore, if $P$ is at least $2r_D$ away from the boundary of $\mathcal W$, the probability of $D$ colliding with $P$ is
\begin{equation}\label{eq:pro}
\Pr=\dfrac{S_{N}}{(H-2r_D)(W-2r_D)}=\dfrac{S_{P}+S_{D}+r_D C_{P}}{(H-2r_D)(W-2r_D)}    
\end{equation}

Assuming both $P$ and $D$ are small compared to $\mathcal W$, $\Pr$ is a fairly tight upper bound of the collision probability of $P$ and $D$ in the workspace.
% Eq.~\ref{eq:pro} suggests that the probability is affected by object area, shape, and relative size to the other objects in the workspace.

% If the size of $P$ changes by a scalar $\alpha$ to $P' (\alpha > 1)$, 
% $$S_{P'}=\alpha S_{P}$$.
% , we have 
% $$\sqrt{\alpha} C_{P} \leq C_{P'} < \alpha C_{P}$$.
% Since $\Pr$ is affected by three terms, depending on the degree of asymmetry of $P$ and the size difference between $P$ and $D$, 
% Therefore, we have:
% \begin{enumerate}
%     \item when $P$ is much larger than $D$, $\Pr'\approx \alpha \Pr$;
%     \item when $P$ is much smaller than $D$, $\Pr'\approx \Pr$; 
%     \item when $P$ and $D$ has similar size, and $P$ has high degree of asymmetry, $\Pr'\approx \Pr$
% \end{enumerate}

Based on Eq.~\ref{eq:pro}, the \heteCP heuristics can be computed as in Algo.~\ref{alg:heteCP}. 
% \heteCP is used to measure the difficulty of buffer allocation.
The algorithm consumes the object characteristics $\mathcal I$. We first compute the area and circumference of each object based on $\mathcal I$ (Lines 1-4).
In Lines 5-6, we create a disc whose size is the average size of workspace objects. 
The area and radius of the average-sized disc are denoted as $\overline{s}$ and $\overline{r}$, respectively.
The weight of each object is then computed as the estimated collision probability between the object and the average-sized disc (Lines 7-9).

In general, \heteCP measures the difficulty of allocating a collision-free pose for an object and the probability of the sampled pose blocking other objects' movements. 

\begin{algorithm}[ht]
%\begin{small}
    \SetKwInOut{Input}{Input}
    \SetKwInOut{Output}{Output}
    \SetKwComment{Comment}{\% }{}
    \caption{\heteCP}
		\label{alg:heteCP}
    \SetAlgoLined
		\vspace{0.5mm}
		\Input{$\mathcal I$: object characteristics.
		}
        \Output{\heteCP: weights on $\mathcal O$}
		\vspace{0.5mm}
    \vspace{1mm}
    \For{$o_i\in \mathcal O$}
    {
    $s_i \leftarrow $ getSize($I_i$)\\
    $c_i \leftarrow $ getCircumference($I_i$)\\
    }	
    \vspace{1mm}
    $\overline{s} = (\sum_{1 \leq i \leq n} s_i)/n$\\
    \vspace{1mm}
    $\overline{r} = \sqrt{\overline{s}/\pi}$\\
    \vspace{1mm}
    \For{$o_i\in \mathcal O$}
    {
    $w_i \leftarrow \dfrac{s_{i}+\overline{s}+\overline{r} c_{i}}{(H-2\overline{r})(W-2\overline{r})}$ \\
    }	
    \vspace{1mm}
    \Return $\{w_1, w_2,\dots, w_n\}$
\end{algorithm}

\subsubsection{Task Impedance (\heteTI) for TI}
When task impedance is evaluated, directly applying the existing methods, which seek rearrangement plans with the minimum number of actions, may lead to undesirable sub-optimality.
For the TI objective, the weighting heuristic is denoted as \heteTI. 
\heteTI gives each object $o_i$ a weight $f(I_i)$ as defined in Eq.~\eqref{eq:ti}.
%\jy{It is confusing to use $w_i$. We should just use $f(I_i)$ or further simplify $f(I_i)$.}

\subsection{Extended TRLB (ETRLB) for \hete}\label{sec:hete-trlb}
Recall that TRLB first computes a primitive plan and then allocates buffers to check plan feasibility.
For \hete, we leverage weighting heuristics to guide the primitive plan computation in TRLB, which yields to \emph{Extended TRLB} (ETRLB).
To do so, we first construct a weighted dependency graph $\mathcal G_w$, where each vertex $v_i$ has a weight $w_i$.
Correspondingly, in the primitive plan computation process, we solve weighted TBM and weighted RBM of $\mathcal G_w$.
For convenience, we use the notations ETBM and ERBM to represent ETRLB with weighted TBM and RBM, respectively.
ETBM computes a primitive plan minimizing the total weights of objects moving to buffers.
We transform the problem into computing the feedback vertex set of $\mathcal G_w$ with the minimum sum of weights, which can be solved by dynamic programming or integer programming similar to TBM in TRLB.
The same as the TRLB with RBM, ERBM computes the minimum running buffer size for the rearrangement problem but it assumes each object $o_i$ to take $w_i$ amount of space in buffers.

In the original RBM computation, TRLB fixes the running buffer size and checks whether there is a feasible rearrangement plan given the running buffer size in order to make use of the efficiency of depth-first search. 
If not, the algorithm increases the running buffer size by one and checks the feasibility again. 
This process continues until it finds the lowest running buffer size with a feasible primitive plan.
Since this framework needs an integer running buffer size, ERBM rounds the weights to the nearest integers so that the sum of weights is still an integer. 

For \hete with the PP objective, applying \heteCP in primitive plan computation discourages relocation of objects for which the allocation of high-quality buffers is difficult, which reduces the challenge in the buffer allocation process.
When weighting heuristic \heteCP is used, ETBM and ERBM seek primitive plans minimizing the total amount of ``difficulty on buffer allocation'' and the maximum ``concurrent difficulty on buffer allocation''.
For the TI objective, we apply \heteTI to primitive plan computation instead.

\subsection{Extended MCTS (EMCTS) for \hete}\label{sec:hete-mcts}
We also propose Extended Monte Carlo Tree Search (EMCTS) for heterogeneous instances.
Specifically, for \hete with the PP objective, EMCTS applies \heteCP to the exploration term in the UCB function, replacing constant $C$ with 
$$C(1+\dfrac{w_i}{\sum^n_{i=1} w_i }).$$
EMCTS prioritizes actions making progress in sending high-weight objects to the goals (moving them to goals or clearing out their goal poses). 
We do this for two reasons. 
First, as mentioned in Sec.~\ref{sec:weighting}, it is difficult to move objects with high \heteCP weights to buffers. 
And their allocated buffers are more likely to block other objects' goal poses.
Second, these high-weighted objects at goal poses are guaranteed to be collision-free from other goal poses.
For \hete with the TI objective, the reward in EMCTS is the total \heteTI weights of objects in the goal poses.

Besides the application of weighting heuristics, to tackle the challenge in \hete, we introduce two additional improvements to both MCTS and EMCTS.
\subsubsection{Reducing action space}
For each state, we only consider actions $a_i$ whose corresponding object $o_i$ are away from goal poses. 
This modification significantly reduces the action space when the state is close to the goal state and will not lead to optimality loss since the neglected actions will not make any difference to the workspace.
    \subsubsection{Efficient collision checker for \hete}
In \toroi, MCTS spends most of the computation time on collision checks for buffer allocation and goal pose feasibility checks.
The original paper\cite{labbe2020monte} uses the bounding discs for collision checks.
For \hete, our collision checker has two steps: in the broad phase, we check collisions between bounding discs of object poses, which aims at pruning out most objects in the workspace that are far from the target object.
In the narrow phase, for objects whose bounding discs overlap with that of the target object, we check the collision of their polygonal approximations.
We also tried other broad-phase collision checkers, such as quadtrees and uniform space partitions. 
While these methods have good performance in large-scale collision checking, we observe limited speed-up but a large amount of overhead in maintaining the structures in \hete instance.
That is because few robotic manipulation instances need to deal with more than 100 objects simultaneously.

\section{Evaluation}\label{sec:evaluation}
The proposed algorithms' performance is evaluated using simulation and real robot experiments.
%\jy{Need to explain the implementation (e.g., the language used) and the machine used to evaluate.}
All methods are implemented in Python, and the experiments are executed on an Intel$^\circledR$ Xeon$^\circledR$ CPU at 3.00GHz. Each data point is the average of $50$ cases except for unfinished trials, given a time limit of $600$ seconds per case. 

\subsection{Numerical Results in Simulation}
\begin{figure}[h]
    \centering
    \vspace{1mm}
    \includegraphics[width=0.48\textwidth]{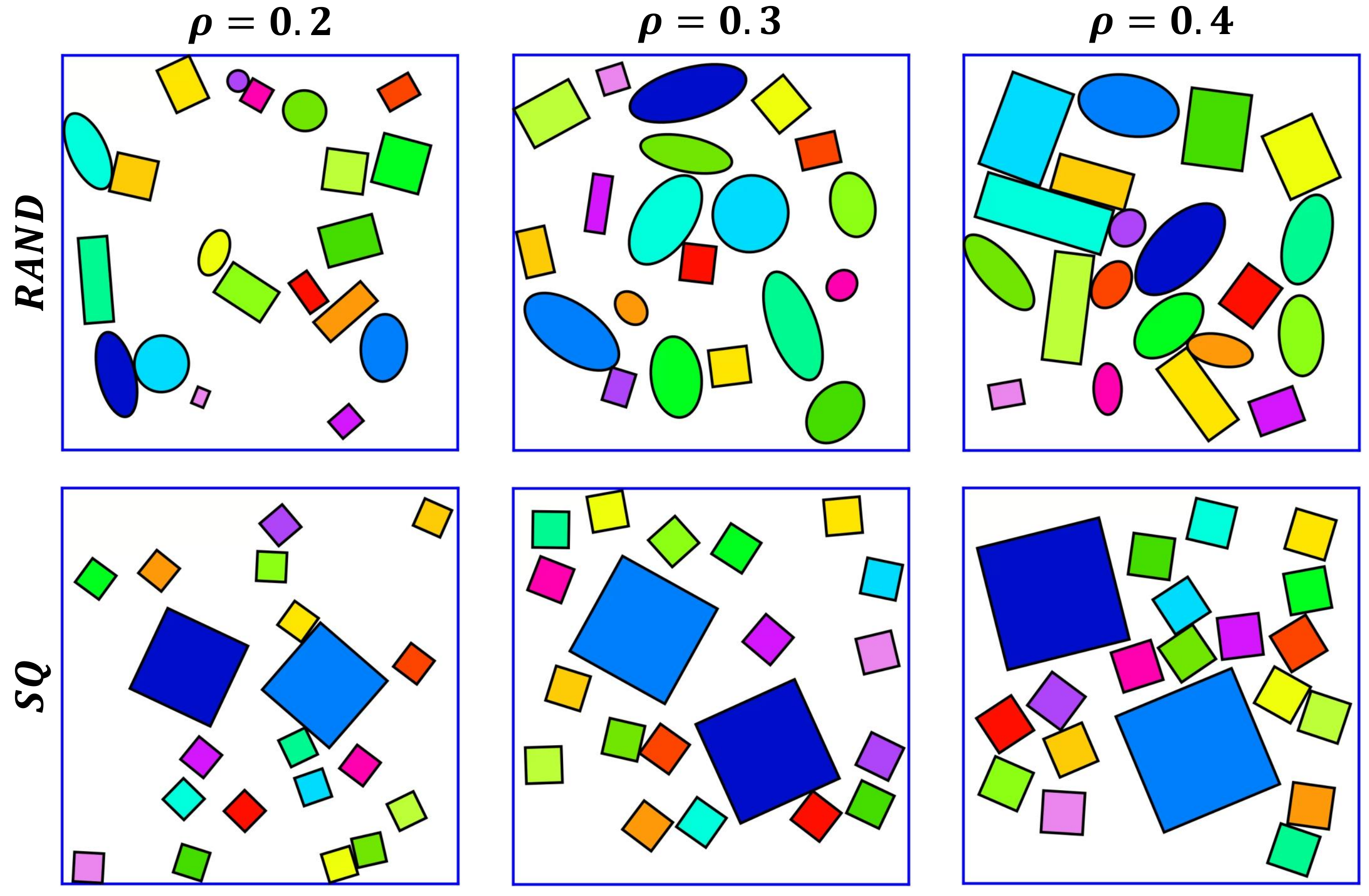}
    \caption{Illustrations of some simulation test cases at different densities. [top] RAND instances under different density levels. [bottom] SQ instances under different density levels.}
    \label{fig:exp_example}
\end{figure}

Simulation experiments are conducted in two scenarios:
\begin{enumerate}[leftmargin=4.5mm]
    \item (RAND) In the RAND (random) scenario, for each object, the base shape is randomly chosen between an ellipse and a rectangle. The aspect ratios and the areas are uniformly random values in $[1.0,3.0]$ and $[0.05,1]$, respectively.
    The areas of the objects are then normalized to a certain sum $S$, whose value depends on specific problem setups. Examples of the RAND scenario are shown in the first row of Fig.~\ref{fig:exp_example}.
    This setup is used to examine algorithm performance in general \hete instances.
    \item (SQ) In the SQ (squares) scenario, there are $2$ large squares and $|\mathcal O|-2$ small squares. The area ratio between the large and small squares is $9:1$. This setup simulates the rearrangement scenario where object size varies significantly.
    Examples of arrangements in the SQ scenario are shown in the second row of Fig.~\ref{fig:exp_example}.
\end{enumerate}

For each test case, object start and goal poses are uniformly sampled in the workspace for both scenarios.
The sampling process for a single test case is repeated in the event of pose collision until a valid configuration is obtained.  
%If the sampled object pose overlaps with existing objects or workspace boundaries, we sample another pose for the object until a collision-free pose is found.
We use the \emph{density level $\rho$} to measure how cluttered a workspace is. 
It is defined as $(\sum_{o_i\in \mathcal O} S_i)/S_W$, where $S_i$ and $S_W$ represents the size of $o_i$ and the workspace respectively. 
In other words, $\rho$ represents the proportion of the workspace occupied by objects. 
Fig.~\ref{fig:exp_example} shows examples of RAND and SQ scenarios for $\rho=0.2$-$0.4$, where $\rho = 0.4$ is already fairly dense (e.g., 3rd column Fig.~\ref{fig:exp_example}).
We limit the maximum number of objects in a test case to $40$ as typical \hete cases are unlikely to have more objects. 

%\jy{For all figures in this section, the fonts are somewhat small. They hurt my eyes a bit. If there is time, please make them one size larger.}

\begin{figure}[h]
    \centering
    \includegraphics[width=0.48\textwidth]{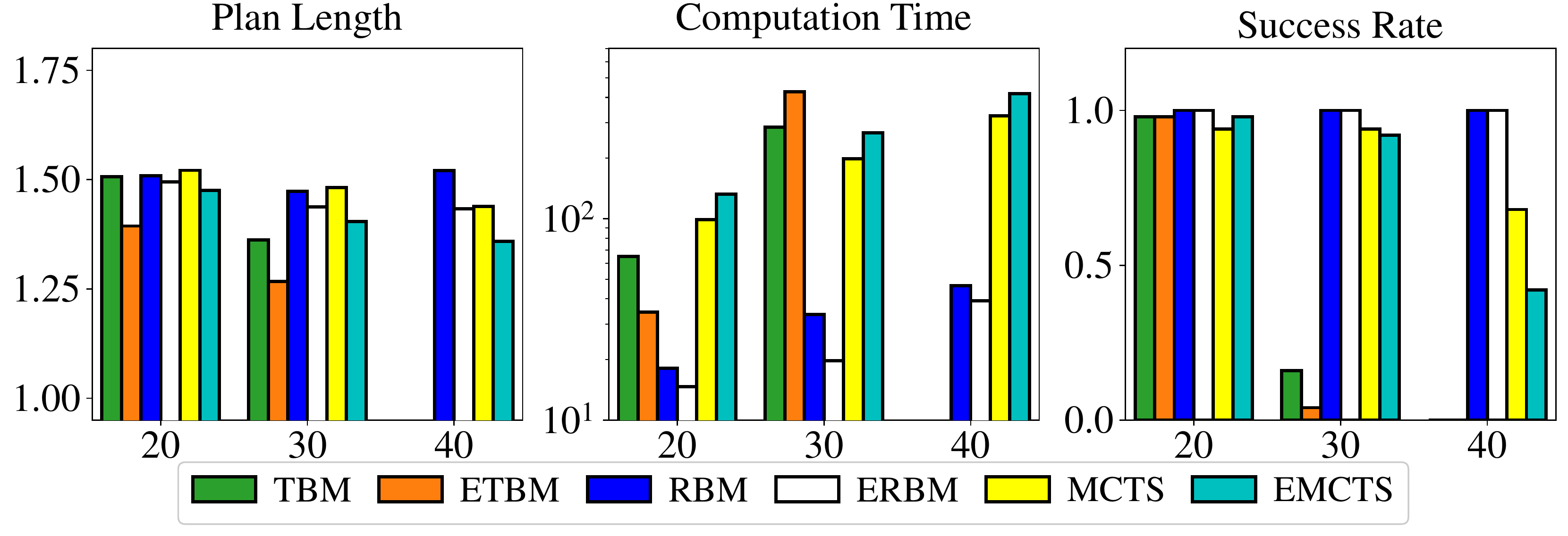}
    \caption{Algorithm performance in RAND instances with $\rho=0.4$ and 20-40 objects. [Left] Plan length as multiplies of $|\mathcal O|$. [Middle] Computation time (secs). [Right] Success rate.}
    \label{fig:rd_pp}
\end{figure}
In Fig.~\ref{fig:rd_pp}, we first evaluate algorithm performance in dense RAND instances ($\rho=0.4$) under the PP objective. 
% In these instances, none of the TBM methods ($TBM$ and $ETBM$) or MCTS methods ($MCTS$ and $EMCTS$) is scalable as the number of objects increases.
Compared with the original TBM, ETBM shows a $7.8\%$ improvement in solution quality with only $52\%$ computation time, when both TBM and ETBM have $98\%$ success rates in 20-object instances.
Suffering from the expansion of solution space, neither TBM or ETBM is scalable as the number of objects increases. 
With \heteCP, both ERBM and EMCTS show clear advantages compared to the original methods (RBM and MCTS) in solution quality as the number of objects increases. 
In 40-object instances, plans computed by ERBM and EMCTS save $3.6$ and $3.2$ actions over those by RBM and MCTS, respectively.
While ERBM spends less time than RBM to solve RAND instances, EMCTS spends slightly more time on computation since the weights are added to the exploration term.
In general, ETBM computes solutions with the best quality in small-scale instances, ERBM is most scalable as $|\mathcal O|$ increases, and EMCTS balances between optimality and scalability,  computing high-quality solutions in 30-object instances with a high success rate.

\begin{figure}[h]
    \centering
    %\vspace{2mm}
    \includegraphics[width=\columnwidth]{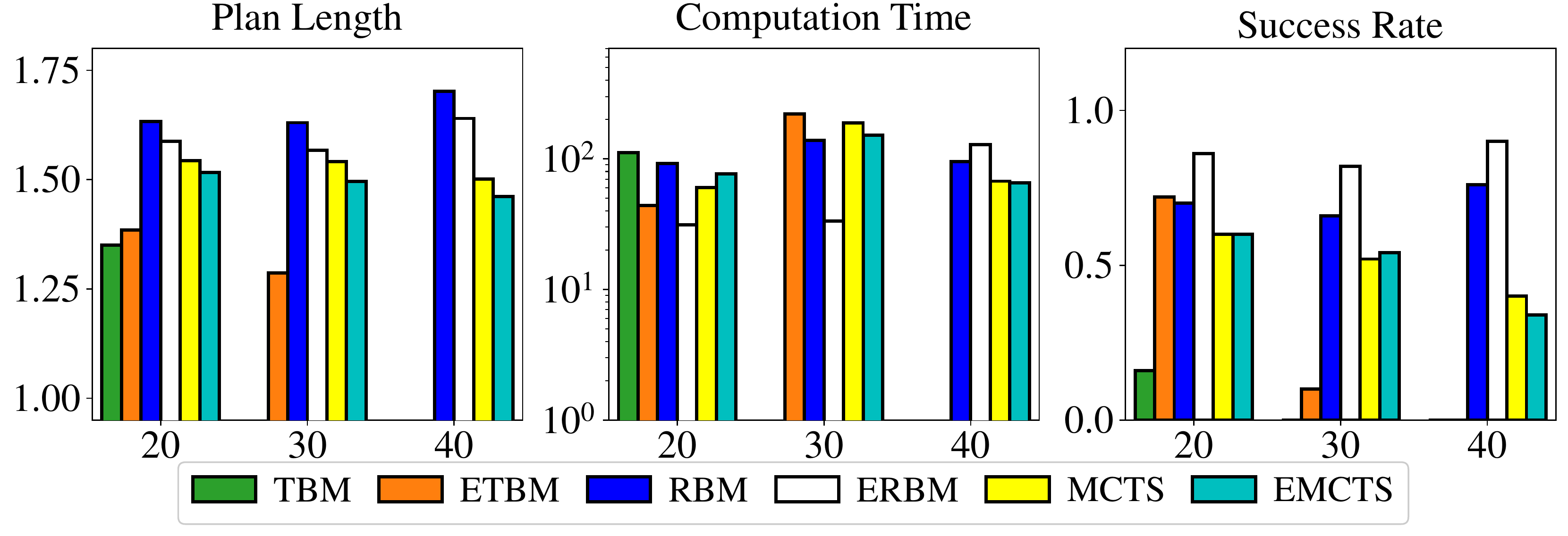}
    \caption{Algorithm performance in SQ instances with $\rho=0.4$ and 20-40 objects. [Left] Plan length as multiplies of $|\mathcal O|$. [Middle] Computation time (secs). [Right] Success rate.}
    \label{fig:sq_pp}
\end{figure}

In SQ instances, it is difficult to displace the large squares in the cluttered workspace. 
Fig.~\ref{fig:sq_pp} shows the algorithm performance in SQ instances with $\rho=0.4$ under the PP objective.
The results suggest that ETBM with \heteCP effectively increases the success rate from $16\%$ to $72\%$ when $|\mathcal O|=20$.
Similarly, the success rate of ERBM is around $15\%$ higher than RBM when $|\mathcal O|=20,30,$ and $40$.
While ERBM solves more test cases, the computed solutions are even $5\%-7\%$ shorter than RBM.
As for MCTS methods, EMCTS computes $3\%-5\%$ shorter paths than EMCTS with a similar success rate and computation time.

\begin{figure}[h]
    \centering
    \includegraphics[width=0.48\textwidth]{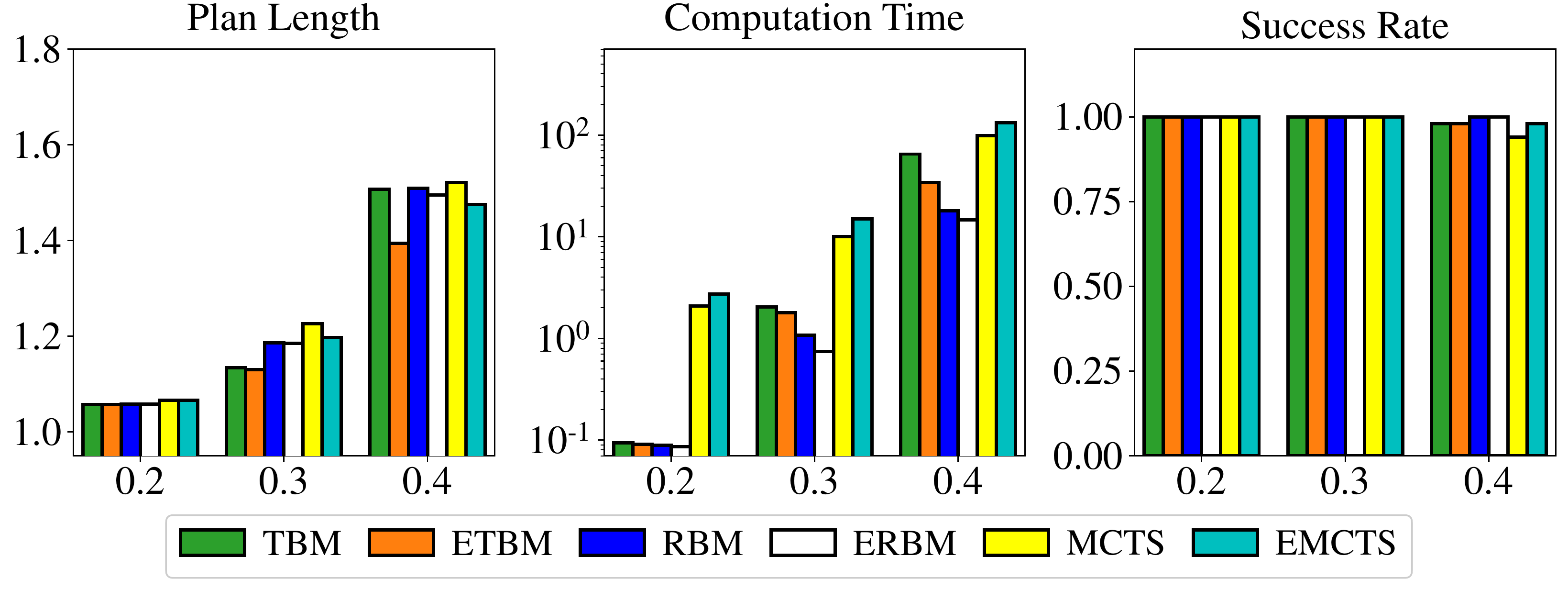}
    \caption{Algorithm performance in RAND instances under different density levels and $|\mathcal O|=20$. [Left] Plan length as multiplies of $|\mathcal O|$. [Middle] Computation time (secs). [Right] Success rate.}
    \label{fig:rd_density}
\end{figure}

\begin{figure}[h]
    \centering
    \vspace{1mm}
    \includegraphics[width=0.48\textwidth]{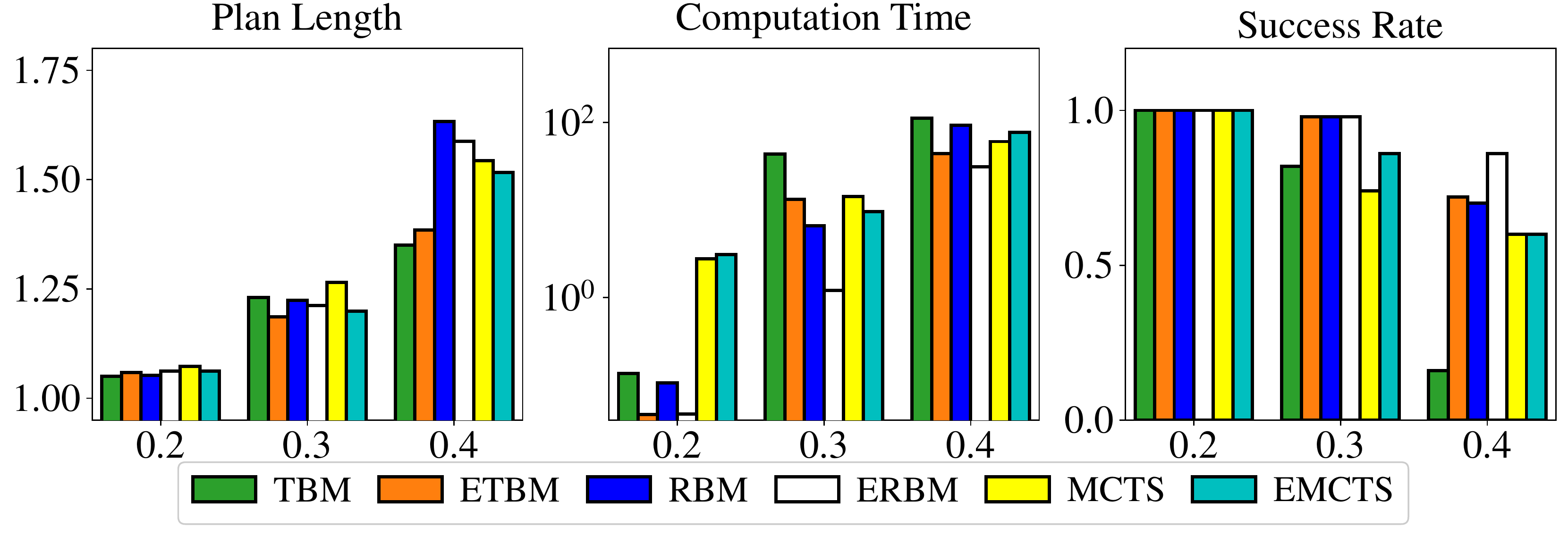}
    \caption{Algorithm performance in SQ instances under different density levels and $|\mathcal O|=20$. [Left] Plan length as multiplies of $|\mathcal O|$. [Middle] Computation time (secs). [Right] Success rate.}
    \label{fig:sq_density}
\end{figure}

Besides dense instances, we also evaluate the efficiency gain of \heteCP at different density levels when $|\mathcal O|=20$.
The results of RAND and SQ scenarios are shown in Fig.~\ref{fig:rd_density} and Fig.~
\ref{fig:sq_density} respectively.
In sparse RAND instances ($\rho=0.2$ or $0.3$), ETBM and ERBM return plans with similar optimality without additional computational overheads.
When $\rho=0.3$, EMCTS spends additional $5$ seconds for exploration and receives around $3\%$ improvements on path length.

In sparse SQ instances, \heteCP helps improve computation time, which contributes to an increase in success rate.
Compared with TBM, ETBM saves $70\%$ computation time when $\rho=0.2$ and $0.3$ and increases the success rate from $82\%$ to $98\%$ when $\rho=0.2$.
Similarly, compared with RBM, ERBM saves $56\%$ and $82\%$ computation time when $\rho=0.2$ and $0.3$.
In ETBM and ERBM, \heteCP speeds up computation without reducing solution quality.
Compared with MCTS, EMCTS computes $5\%$ shorter plans, saves $33\%$ computation time, and increases the success rate by $16\%$ when $\rho=0.3$.

\begin{figure}[h]
    \centering
    \vspace{2mm}
    \includegraphics[width=0.48\textwidth]{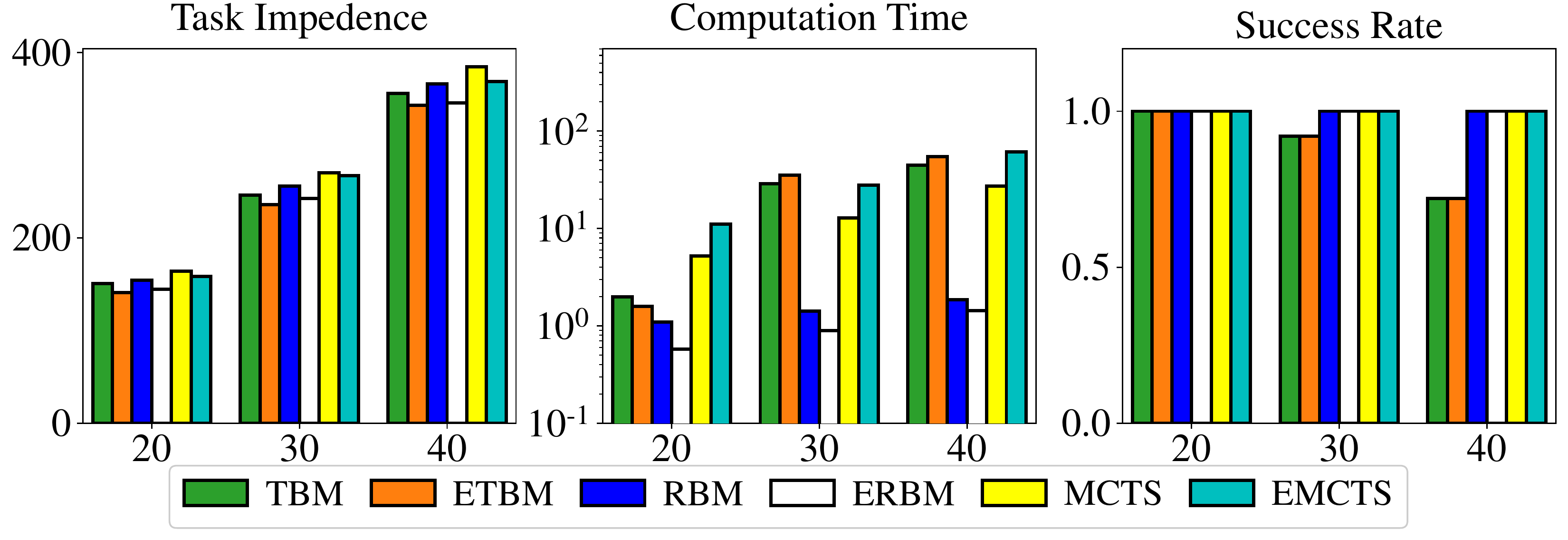}
    \caption{
    Algorithm performance in RAND instances 
    with TI objective. $\rho=0.3$, $20 \leq |\mathcal O|\leq 40$.
    [Left] Total task impedance. [Middle] Computation time (secs). [Right] Success rate.}
    \label{fig:rand_effort}
\end{figure}

\begin{figure}[h]
    \centering
    \includegraphics[width=0.48\textwidth]{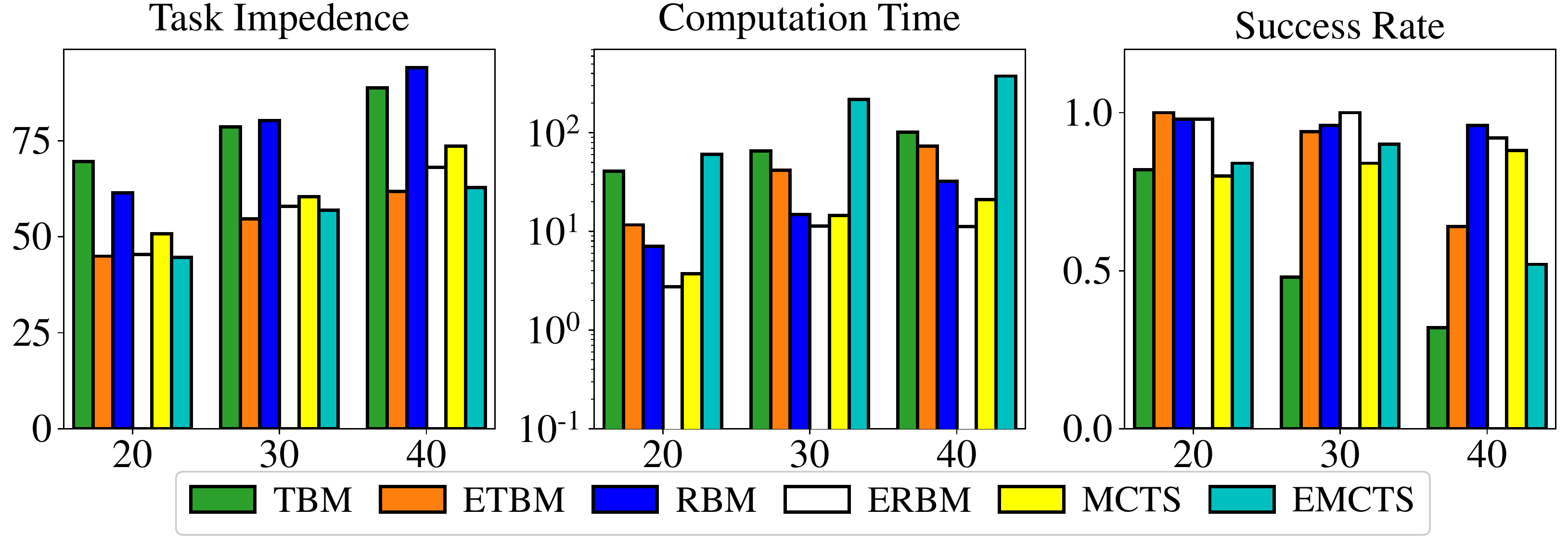}
    \caption{Algorithm performance in SQ instances 
    with TI objective. $\rho=0.3$, $20 \leq |\mathcal O|\leq 40$.
    [Left] Total task impedance. [Middle] Computation time (secs). [Right] Success rate.}
    \label{fig:squares_effort}
\end{figure}

In Fig.~\ref{fig:rand_effort} and Fig.~\ref{fig:squares_effort}, we show the algorithm performance of \hete with TI objective in RAND and SQ instances respectively.
$\rho=0.3$ in the test cases.
In RAND instances, where only around $10\%$ objects need buffer locations, the total task impedance of computed plans is similar.
Compared with original methods (TBM, RBM, and MCTS), ETBM, ERBM, and EMCTS reduce task impedance by $1\%-7\%$.
In SQ instances, object impedance varies greatly. In this case, methods with \heteTI compute much better solutions.
Compared with TBM, ETBM reduce task impedance by over $30\%$.
Similarly, ERBM reduces task impedance by around $27\%$.
Compared with MCTS, EMCTS reduces task impedance by $6\%-15\%$ but with additional overhead.

\subsection{Demonstrations in Simulation and Real-World}
\begin{figure}[h]
    \centering
    \vspace{1mm}
    \includegraphics[width=\columnwidth]{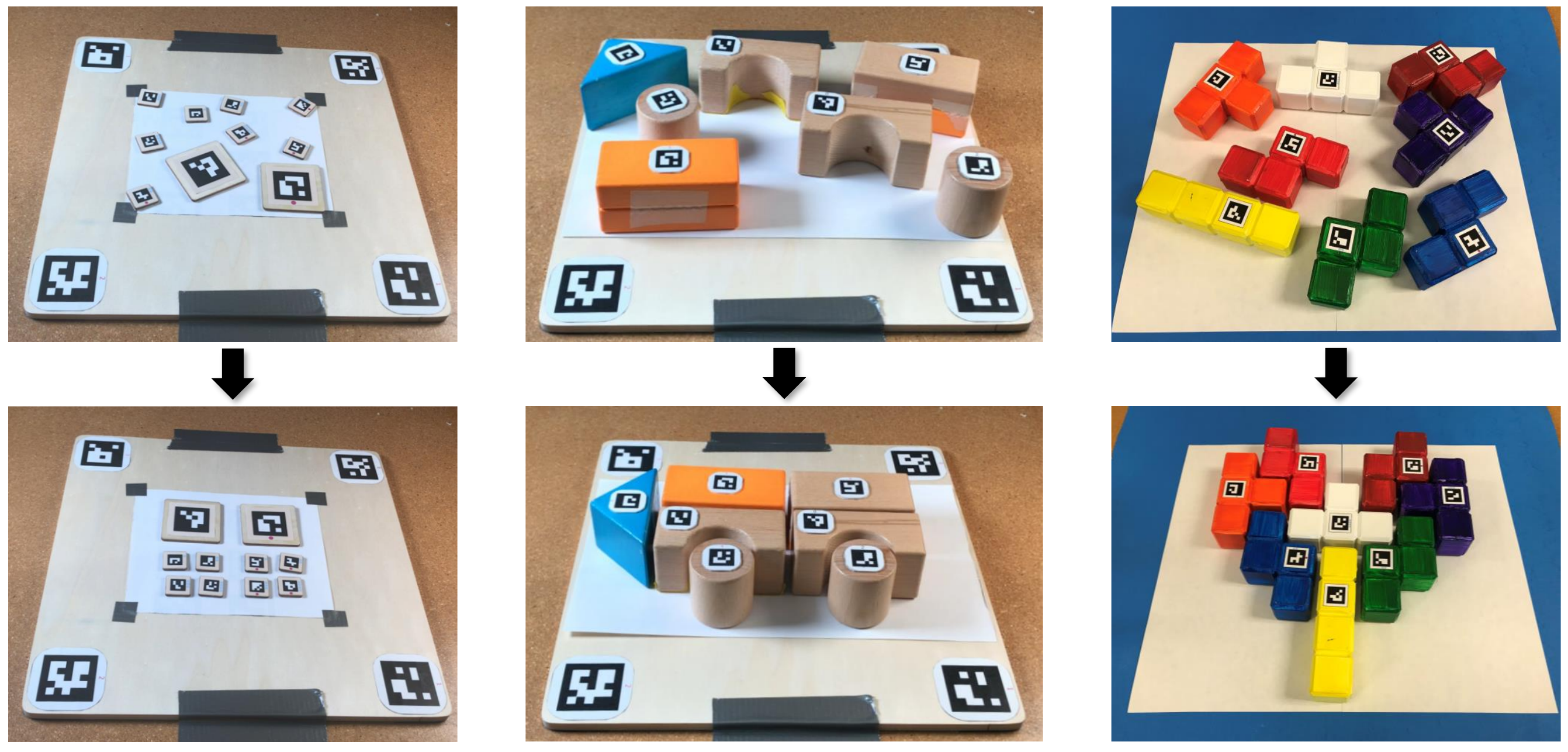}
    \caption{Physical experiment/demo setups. [left] Dense SQ instance with $10$ square wooden planks. [middle] Rearranging densely-packed thick wooden blocks of different shapes and sizes. [right] Rearranging 3d-printed Tetris pieces into a heart shape.}
    \label{fig:demo}

\end{figure}

In the accompanied video, we further demonstrate that our algorithms can be applied to practical instances with general-shaped objects in both simulation and the real world. 
For simulation instances, the rearrangement plans are executed in Pybullet. 
For real-world instances, the objects are detected with an Intel RealSense D405 RGB-D camera and manipulated with an OnRobot VGC 10 vacuum gripper on a UR-5e robot arm. Since object recognition and localization are not the focus of this study, we use fiducial markers for realizing these. 
The workspaces for rearrangement are rather confined (in Fig.~\ref{fig:demo}, the white areas for the first two and the wooden board for the third), translating to very high object densities, between $35$-$37\%$.
In the SQ instance with 10 wooden planks, our proposed methods ETBM and ERBM offer advantages over TBM and RBM in both solution quality and computation time, saving $1$ and $4$ actions, respectively. 
Furthermore, EMCTS is able to find a $15$-action plan in just $13.70$ secs while MCTS fails to find a solution even after $2$ minutes of computation.
%\jy{The $30$-$50\%$ is just my guess. We should replace them with actual numbers.}

%\jy{Fig.~\ref{fig:demo} can be larger. We should have space.}
%\justin{I agree that it should be larger. I think it might help if we have the "before" pictures at the top row and point the arrows all down to the bottom row for the "goal" pictures.}
%\jy{Maybe we don't have YCB. As Justin suggested, we can have the three real cases arranged in a top-down manner so we have three in a row, which will be larger.}

\section{Conclusions}\label{sec:conclusion}
In this work, we examine rearranging heterogeneous objects on a tabletop (\hete) using two different objectives: PP and TI. 
The PP objective assumes each object requires an equal amount of manipulation cost for a pick-n-place, while utilizing the object's shape and size can enhance rearrangement efficiency. 
On the other hand, the TI objective evaluates the total task impedance, where the manipulation cost of each object varies based on its specific characteristics, such as size or weight. 
To optimize these objectives, we propose weighting heuristics and build SOTA algorithms with them. Extensive simulation experiments confirm the efficiency of our proposed methods. 
Moreover, we demonstrate the effectiveness of our methods in challenging real-world \hete instances through multiple demonstrations on a physical robot platform.

% \newpage
{\small
\bibliographystyle{formatting/IEEEtran}
\bibliography{bib/bib}
}

\end{document}